\documentclass{article}

\PassOptionsToPackage{numbers, compress}{natbib}

\usepackage[preprint]{neurips_2023}

\usepackage[utf8]{inputenc} 
\usepackage[T1]{fontenc}   
\usepackage{hyperref}      
\usepackage{url}            
\usepackage{booktabs}       
\usepackage{amsfonts}       
\usepackage{nicefrac}       
\usepackage{microtype}      
\usepackage{xcolor}        
\usepackage{amsmath}
\usepackage{tcolorbox}
\tcbuselibrary{breakable}
\tcbset{breakable}
\usepackage{makecell} 
\usepackage{array}    
\usepackage{float}
\usepackage{graphicx}
\usepackage{ltablex}
\usepackage{lipsum}
\usepackage{graphicx}
\usepackage{wrapfig}
\usepackage[size=small]{caption}
\usepackage{mathtools}
\usepackage{tabularx}
\usepackage{amssymb}
\usepackage{amsthm}
\usepackage{soul}

\definecolor{textblue}{rgb}{.2,.2,.7}
\definecolor{textred}{rgb}{0.54,0,0}
\definecolor{textblack}{rgb}{0,0,0}
\definecolor{textgreen}{rgb}{0,0.43,0}
\usepackage{listings}
\lstdefinestyle{pythonstyle}{
    language=Python,
    morekeywords={None}, 
    breaklines=true,
    basicstyle=\ttfamily\small,
}

\title{Octopus v2: On-device language model for super agent}

\author{Wei Chen$^{\dagger}$ \thanks{Corresponding author, $^\dagger$ equal contribution} \\
Stanford University\\
\texttt{\{weichen6\}@stanford.edu} \\
\And
Zhiyuan Li$^\dagger$ \\
Stanford University\\
\texttt{\{zhiyuan8\}@stanford.edu} \\
}

\begin{document}

\maketitle

\begin{abstract}
Language models have shown effectiveness in a variety of software applications, particularly in tasks related to workflow automation. These models possess the crucial ability to call functions, which is essential for creating AI agents. Despite the high performance of large-scale language models in cloud environments, they are often associated with concerns over privacy and cost. Current on-device models for function calling face issues with latency and accuracy. Our research presents a new method that empowers an on-device model with 2 billion parameters to surpass GPT-4 in both accuracy and latency, while reducing the context length by 95\%. Compared to Llama-7B with a RAG-based function calling mechanism, our method improves latency by 35-fold. This method reduces latency to levels deemed suitable for deployment across a variety of edge devices in production environments, aligning with the performance requirements of real-world applications.

\end{abstract}

\begin{figure}[h]
    \centering
    \includegraphics[width=0.75\textwidth]{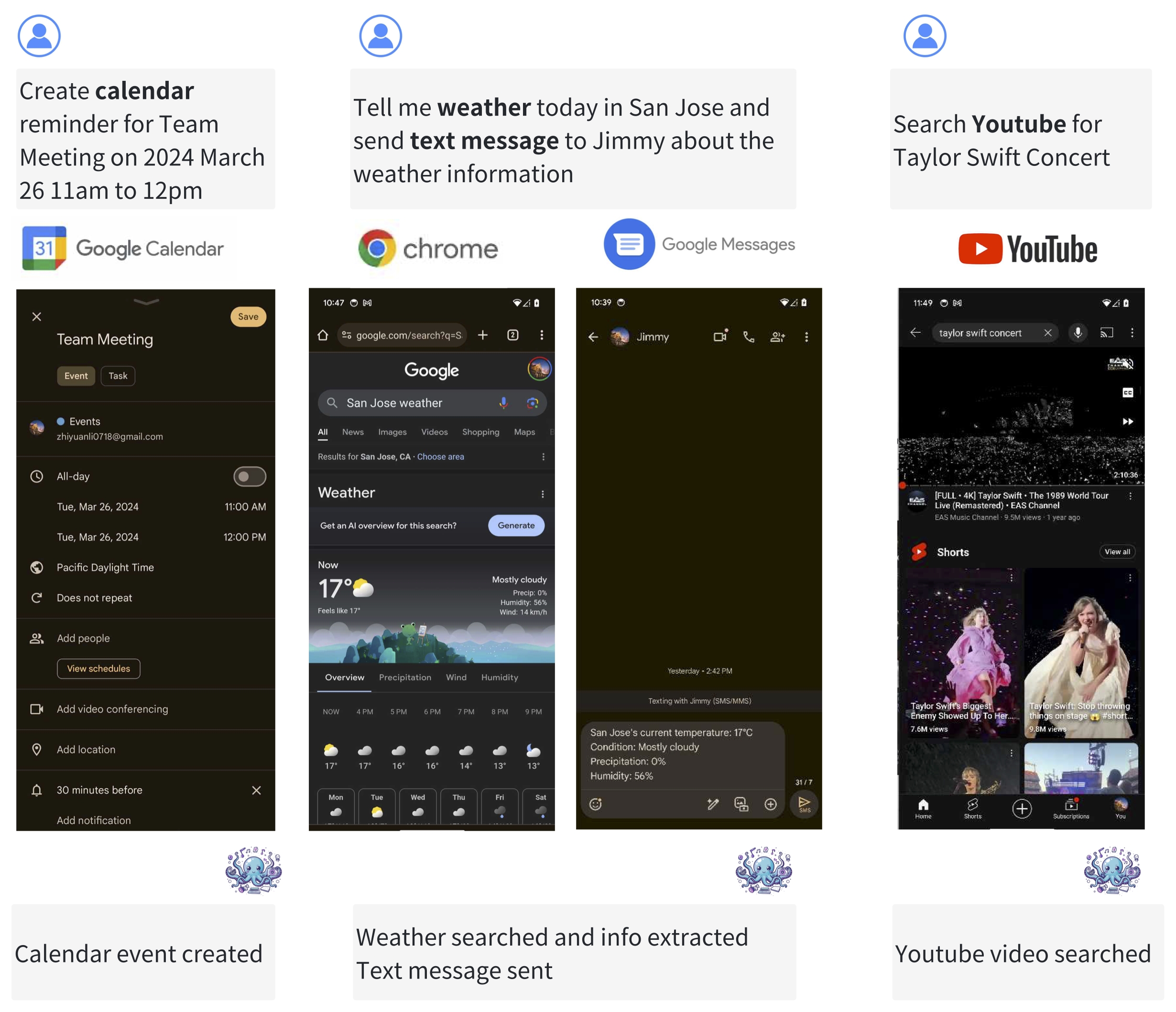}
    \caption{Automated smartphone workflow using the Octopus model.}
    \label{fig:tool-usage}
\end{figure}

\section{Introduction}
Large language models have demonstrated impressive capabilities in function calling, significantly contributing to AI agents' growing presence in the software industry \cite{wang2023survey,brynjolfsson2023generative,hauptman2023adapt,dong2023towards,du2024anytool}. The advancement of AI agents is rapid, highlighted by AI assistant tools such as MultiOn \cite{multionai2024} and Adept AI \cite{adeptai2024}, and AI consumer products such as the Rabbit R1 \cite{rabbit2024} and the Humane AI Pin \cite{humane2024}, which are gaining traction in the consumer sector. Research into AI agents has been robust, with developments in chain-of-thought reasoning \cite{zhang2022automatic, jie2023design} and enhanced prompting techniques \cite{wei2022chain}. Moreover, the rise of multi-agent systems \cite{wu2023autogen, talebirad2023multi, shen2024hugginggpt,paranjape2023art} marks a novel trend in the industry, showcasing the use of language models to develop dependable software that empowers users \cite{xi2023rise,shen2024small}. These innovations leverage the API calling \cite{yang2023appagent,hong2023metagpt,wang2024mobile} and reasoning abilities \cite{shinn2023reflexion,ruan2023tptu} of large, cloud-based language models to convert human natural language instructions into actionable commands. Despite the considerable progress in creating valuable AI agents, reliance on cloud models raises issues concerning privacy, inference costs, and the need for Wi-Fi connectivity \cite{yao2024survey,liu2023prompt}.

The cost of using large language models such as Google's Gemini family models \cite{team2023gemini} and OpenAI's GPT series models \cite{radford2018improving, radford2019language, brown2020language, achiam2023gpt} can be substantial; for example, an hour-long interaction with an AI bot might cost around 0.24 USD according to the GPT-4 API price. For function calling, employing RAG-based \cite{lewis2020retrieval,mao2020generation,li2022survey, jiang2023active} or context-augmented \cite{ram2023context} methods requires processing about 1000 tokens for each call, resulting in costs of approximately 0.01 USD. In practical applications, where hundreds of function calls may be made, the cumulative cost can be considerable. Additionally, the potential for privacy violations deters many from using GPT-4, amid concerns that sensitive information might be exposed.

To mitigate costs and enhance privacy, there is a trend towards creating smaller models for deployment on edge devices like smartphones, cars, VR headsets, and personal computers \cite{wen2023empowering,dettmers2024qlora,lin2023awq,li2023mimic,xu2023llmcad,hong20233d,spector2023accelerating}. However, models deployed on edge devices face challenges with high latency, leaving them far from production readiness, and the limited battery life of edge devices further complicates continuous interaction. Research shows that energy consumption reaches 0.1J per token for 1 billion parameter models \cite{liu2024mobilellm}. Therefore, employing a 7B parameter model for function calls with traditional retrieval-augmented methods would consume 700J per call, roughly 1.4\% of a 50kJ iPhone battery, limiting the device to around 71 function calls.

Smaller models often fall short in reasoning tasks and demand extensive tuning for effective function calling. To address these issues, we developed a method that enhances both accuracy and latency for function calling in 2B parameter on-device models, achieving state-of-the-art (SOTA) results. This approach involves tokenizing the core function's name and fine-tuning the model with \textit{functional tokens}. Fine-tuning with these tokens allows the model to understand software application capabilities through additional special tokens, learning to map function descriptions to specific tokens. In the inference phase, the model uses \textit{functional tokens} to achieve better function calling performance than GPT-4. We present a 2B parameter model fine-tuned from Gemma 2B \cite{gemma-2023-open-models}, saving over 95\% context length during model inference. For iPhone use, this enables 37 times more function calls with the same battery and reduces latency by approximately 35 times per function call.

\section{Related work}

\textbf{Deployment of on-device language models} \quad Due to memory limitations and lower inference speeds, deploying larger models on edge devices like PCs or smartphones is challenging. Nonetheless, efforts to deploy smaller-scale Large Language Models (LLMs) to edge devices are underway. Open-source models of manageable sizes, such as Gemma-2B, Gemma-7B, StableCode-3B \cite{stable-code-3b}, and Llama-7B \cite{touvron2023llama}, have been introduced. To enhance these models' on-device inference speed, research initiatives such as llama.cpp \cite{llama-cpp} have been developed. The MLC LLM framework \cite{mlc-llm} allows the operation of 7B language models on mobile phones and other edge devices, demonstrating compatibility across various hardware, including AMD, NVIDIA, Apple, and Intel GPUs.

\textbf{Function calling in language models} \quad Rapid advancements have been observed in the function-calling capabilities of smaller-scale models. Projects such as NexusRaven \cite{srinivasan2023nexusraven}, Toolformer \cite{schick2024toolformer}, ToolAlpaca \cite{tang2023toolalpaca}, Gorilla \cite{patil2023gorilla}, ToolLlama \cite{qin2023toolllm}, and TaskMatrix \cite{liang2023taskmatrix} have demonstrated that 7B and 13B models can call external APIs with efficacy comparable to GPT-4. The pioneering Octopus v1 project even enabled a 2B model to perform on par with GPT-4. This body of work utilizes a RAG-based method for function calling, in which the model retrieves relevant functions from a large pool based on the user's query and then generates a response using these functions as context.

\textbf{Fine-tuning and adaptors of language models} \quad Fine-tuning language models has become a prevalent practice, with various efforts dedicated to this endeavor. LoRA \cite{hu2021lora} is often the method of choice for training models under GPU resource constraints. We use both full model training and LoRA training in our work, and compare their performance. A notable benefit of LoRA is its facilitation of extended functionalities in models, suggesting its potential to adapt our current framework for a broad range of applications.

\section{Methodology}
In this section, we detail the primary methodology implemented in our models, followed by the dataset collection process essential for fine-tuning these models. We illustrate this through examples drawn from the Android API. Subsequently, we delve into the specifics of our model training approach.

\subsection{Causal language model as a classification model}
To successfully invoke a function, it is essential to accurately select the appropriate function from all available options and to generate the correct function parameters. This entails a two-stage process: a function selection stage and a parameter generation stage. The initial step involves understanding the function's description and its arguments, using information from the user's query to create parameters for an executable function. A direct strategy might combine a classification model with a causal language model. We can envision the $N$ available functions as a selection pool, transforming the selection challenge into a softmax classification problem.

One straightforward method for classification is retrieval-based document selection, which identifies the function that most closely matches the user's query by semantic similarity; another is to use a classification model to map the query to a specific function name. Alternatively, autoregressive models, such as a GPT model, can predict the correct function name from the user's query within the context of potential functions. Both approaches essentially divide the task into two parts, potentially requiring two models, $\pi_1$ and $\pi_2$:
\begin{equation}
P(f|q) = P(f|q;\pi_1), \quad P(\operatorname{params}|f,q) = P(\operatorname{params}|f,q;\pi_2),
\end{equation}
where $q$ denotes the query, $f$ denotes the selected function name, and \textit{params} denotes the parameters for the chosen function. Motivated by the principles of multitask learning/meta-learning \cite{caruana1997multitask}, and to achieve faster inference and system convenience, we pursue a unified GPT model strategy, setting $\pi_1=\pi_2=\pi$. Therefore, we redefine our objective as:
\begin{equation}
P(f, \operatorname{params}|q) = P(f|q;\pi) P(\operatorname{params}|f, q; \pi).
\end{equation}

For $P(f|q;\pi)$, the traditional method involves retrieving relevant functions and providing context about several pertinent functions to deduce the optimal function name. In most use cases, the set of possible function names is fixed. When utilizing a language model to formulate a function name, multiple tokens must be generated to form a single function name, which can lead to inaccuracies. To mitigate such errors, we propose designating functions as unique \textit{functional tokens}. For example, in a pool of $N$ available functions, we assign token names ranging from \texttt{<nexa\_0>} to \texttt{<nexa\_{N-1}>} to symbolize these functions. This transforms the prediction task for function names into a single-token classification among the $N$ functional tokens, enhancing the accuracy of function name prediction while simultaneously reducing the number of tokens required. To implement this, we introduce new special tokens from \texttt{<nexa\_0>} to \texttt{<nexa\_{N-1}>} into the tokenizer and modify the architecture of the pretrained model to expand the language head by an additional $N$ units. Thus, for function name prediction, we utilize the language model to pinpoint the correct function among the $N$ functional tokens through argmax probability selection.

To choose the correct \textit{functional token}, the language model must grasp the meaning associated with that token. We therefore incorporate the function descriptions into the training dataset, enabling the model to learn the meaning of these specialized tokens. We designed a prompt template that accommodates three different response styles, facilitating parallel and nested function calls. Detailed examples of the dataset are provided in the Appendix.
\begin{tcolorbox}
Below is the query from the users, please choose the correct function and generate the parameters to call the function.

Query: \texttt{\{query\}}\\

\# for single function call

Response: \texttt{<nexa\_i>(param1, param2, ...)<nexa\_end>}\\

\# for parallel function call

Response: \texttt{<nexa\_i>(param1, param2, ...);\texttt{<nexa\_j>(param1, param2, ...)}<nexa\_end>}\\

\# for nested function call

Response: \texttt{<nexa\_i>(param1, \texttt{<nexa\_j>(param1, param2, ...)}, ...)<nexa\_end>}

Function description: 
\{function\_description\}
\end{tcolorbox}





\begin{figure}[h]
    \centering
    \includegraphics[width=1.0\textwidth]{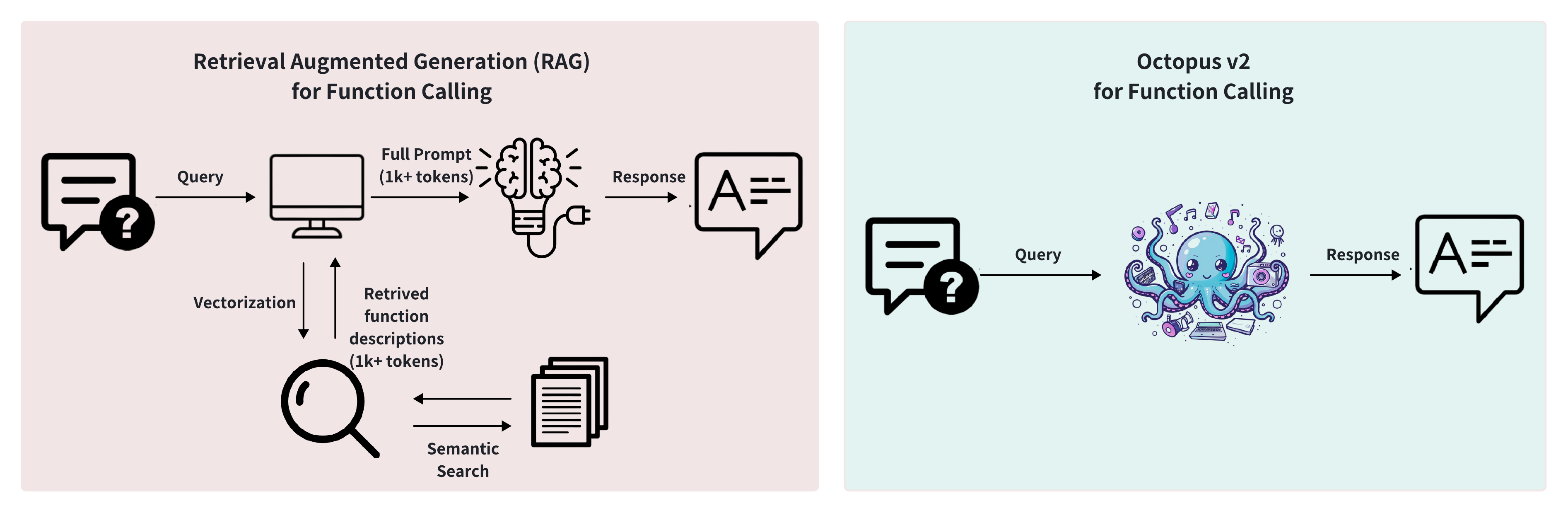}
    \caption{Comparison between the retrieval-based function calling process and the function calling process of the Octopus model.}
    \label{fig:diff}
\end{figure}

This methodology offers an additional critical benefit. Once the model is fine-tuned to understand the significance of functional tokens, it can conduct inference by employing the added special token, \texttt{<nexa\_end>}, as the \textbf{early stopping criterion}. This strategy eliminates the need to analyze tokens from function descriptions, removing both the retrieval of relevant functions and the processing of their descriptions. Consequently, it considerably reduces the number of tokens needed to accurately identify a function name. The difference between the conventional retrieval-based method and our proposed model is shown in Figure (\ref{fig:diff}).

\subsection{Functional token}
The critical innovation presented in this paper is the introduction of the functional token. Drawing parallels to the tokenization of natural language, we propose representing specific functions as functional tokens. We introduce a training methodology for these tokens, inspired by techniques used in natural language models for handling rare words, particularly the word2vec \cite{mikolov2013efficient} framework, in which contextual words enrich a token's semantic representation. For instance, pretrained language models may initially struggle to recognize specialized terms such as \textit{PEGylation} and \textit{Endosomal Escape} from the domain of chemistry. However, these models can learn such terms through causal language modeling, leveraging corpora that include these specialized terms. Similarly, functional tokens can be learned using comparable methods. Our experiments have not identified any limitation on the number of functional tokens that can be defined, allowing users to map any specific function to a token.

By incorporating functional tokens, we aim to assign them the same importance as other linguistic tokens. Unlike conventional linguistic tokens, our functional token does not possess inherent natural language meaning; instead, it represents specific actions encapsulated within the model. For smaller-scale models such as Google Gemma-2B, function calling can be challenging, although they excel at typical language completion tasks. The functional token enables the language model to transform function calling into a standard completion task. As we define the actions for the language model, we can also characterize our model as a \textit{small action model}. The integration of functional tokens allows the model to concentrate on a fixed set of actions and perform effectively on these tasks.

\subsection{Dataset collection}
This section outlines our methodology for assembling high-quality datasets for training, validation, and testing. It also describes the organized process we used to structure the dataset for efficient training.

\textbf{API Collection} \quad As an example, we start with Android APIs. Our selection criteria encompass usability, usage frequency, and the complexity of technical implementation. We ultimately gather 20 Android APIs and organize them into three separate categories, ensuring that each function can be realistically executed on devices through Android app development, provided the developer possesses the necessary system permissions. We also compile APIs available in vehicles. More examples can be found in the Appendix.

\begin{enumerate}
    \itemsep0em
    \parsep0em
    \item \textbf{Android system API} \quad This category includes APIs for system-level functions essential for basic mobile operations, such as making calls, texting, setting alarms, modifying screen brightness, creating calendar entries, managing Bluetooth, enabling do-not-disturb mode, and taking photos. We exclude highly sensitive tasks like accessing system state information or changing accessibility settings.
    \item \textbf{Android App API} \quad Our research examines APIs from pre-installed Google apps on Android devices, such as YouTube, Google Chrome, Gmail, and Google Maps. We explore functionalities like accessing trending news, retrieving weather updates, searching for YouTube content, and map navigation.
    \item \textbf{Android smart device management API} \quad Our focus extends to the Google Home ecosystem, which comprises a wide range of smart home devices with significant market presence. Our aim is to improve smart device management via APIs, covering functions like adjusting a Nest Thermostat, managing media playback on a Google Nest device, and controlling door locks using the Google Home App.
\end{enumerate}

\begin{figure}[h]
    \centering
    \includegraphics[width=1.0\textwidth]{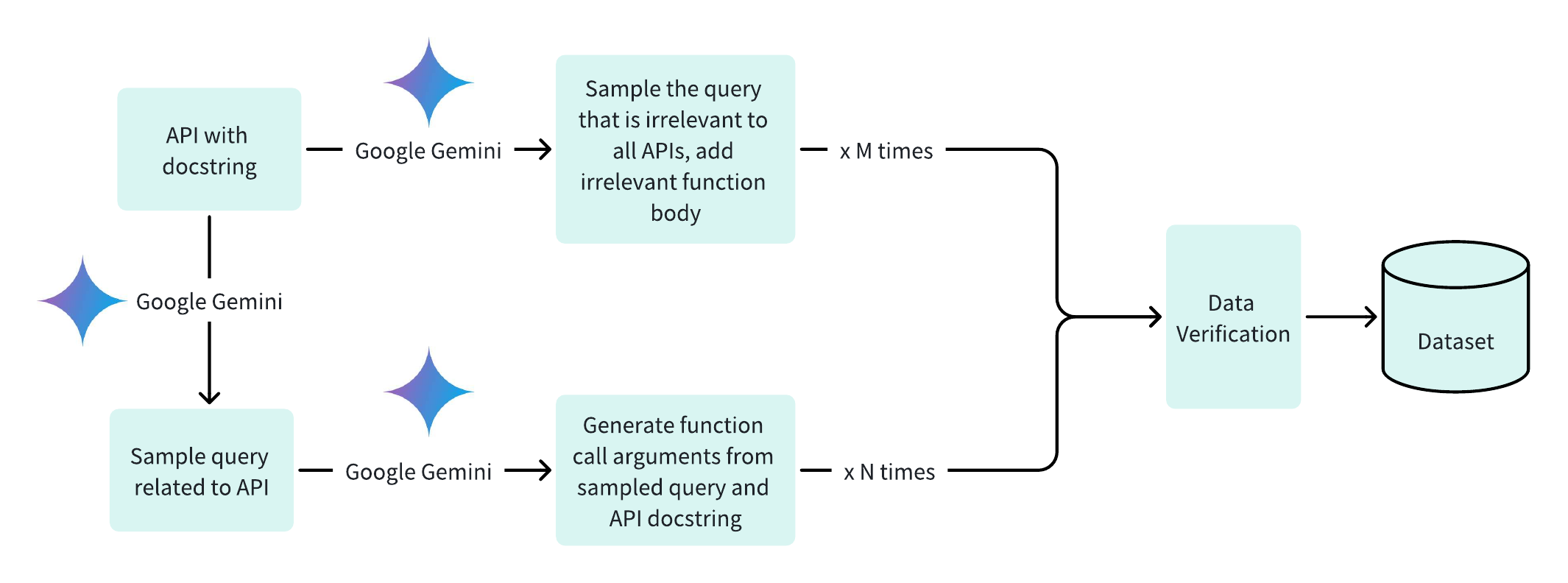}
    \caption{The dataset generation process involves two critical stages: (1) the creation of solvable queries specific to certain APIs, together with the generation of appropriate function calls for them, and (2) the creation of unsolvable queries, complemented by unrelated function bodies. A binary validation mechanism provides rigorous verification, ensuring the collection of an optimized training dataset poised to significantly improve model functionality.}
    \label{fig:data-pipeline}
\end{figure}

\textbf{Dataset generation} \quad Our approach is depicted in Figure (\ref{fig:data-pipeline}), which shows the steps involved in assembling the dataset. The creation of the dataset involves three key phases: (1) generating relevant queries and their associated function call arguments; (2) developing irrelevant queries accompanied by suitable function bodies; and (3) implementing binary verification support through Google Gemini.

\begin{enumerate}
    \itemsep0em
    \parsep0em
    \item \textbf{Google Gemini Generated Query and Function Call} \quad Creating a high-quality dataset hinges on formulating well-defined queries and accurate function call arguments. Our strategy emphasizes generating positive queries that a single API can resolve. With a query and predetermined API descriptions in hand, we utilize a subsequent Google Gemini API call to produce the required function call arguments.
    \item \textbf{Negative Samples} \quad To enhance the model's analytical skills and practical applicability, we incorporate examples from both positive and negative datasets. The balance between these sets, represented by the ratio $\frac{M}{N}$ in Figure \ref{fig:data-pipeline}, is fundamental to our experimental methodology. Specifically, we select $M$ and $N$ to be equal, each assigned a value of 1000.
\end{enumerate}

\textbf{Dataset Verification}\quad Despite the advanced capabilities of large language models such as OpenAI's GPT-4 and Google's Gemini, there remains a noticeable rate of errors, particularly in the generation of function call arguments. These errors may manifest as missing arguments, incorrect argument types, or misinterpretations of the intended query. To mitigate these shortcomings, we introduce a verification mechanism. This system allows Google Gemini to evaluate the completeness and accuracy of its generated function calls and, should the output be found lacking, to initiate a regeneration process.

\subsection{Model development and training}
We employ the Google Gemma-2B model as the pretrained model in our framework. Our approach incorporates two distinct training methodologies: full model training and LoRA training. For full model training, we utilize the AdamW optimizer with a learning rate of \texttt{5e-5}, 10 warm-up steps, and a linear learning rate scheduler. The same optimizer and learning rate configuration is applied to LoRA training. We set the LoRA rank to 16 and apply LoRA to the following modules: \texttt{q\_proj}, \texttt{k\_proj}, \texttt{v\_proj}, \texttt{o\_proj}, \texttt{up\_proj}, and \texttt{down\_proj}. The LoRA alpha parameter is set to 32. For both training methods—full model and LoRA—we set the number of epochs to 3.

\begin{table}[ht]
\centering
\caption{Benchmark accuracy and per-call latency on the Android function-call evaluation set. The Octopus variants differ only in training dataset size and training method; their configurations are given in Table~\ref{tab:octopus_models}. The best result in each column is shown in bold.}
\label{tab:benchmark}
\vspace{4pt}
\setlength{\tabcolsep}{14pt}
\begin{tabular}{@{}lcc@{}}
\toprule
Model & Accuracy (\%) & Latency (s) \\
\midrule
Llama-7B + RAG & 68.095 & 13.46 \\
GPT-3.5 + RAG  & 98.095 & 1.97 \\
GPT-3.5        & 97.143 & 1.18 \\
GPT-4          & 98.571 & 1.02 \\
\midrule
Octopus-0      & \textbf{99.524} & 0.38 \\
Octopus-1      & 99.048 & 0.37 \\
Octopus-2      & 99.048 & \textbf{0.36} \\
Octopus-3      & 98.095 & 0.38 \\
\bottomrule
\end{tabular}
\end{table}

\section{Experiments}
Our study conducts a comprehensive evaluation of language model capabilities through an extensive benchmarking approach, aimed at assessing their effectiveness in generating accurate function calls. Initially, we compare our model's accuracy and response time against leading models in the field, namely GPT-4 (checkpoint: \textit{gpt-4-0125-preview}) and GPT-3.5 (checkpoint: \textit{gpt-3.5-turbo-0125}).

In the next phase, we explore the efficacy of the RAG technique, renowned for its ability to reduce incorrect outputs (hallucinations) and latency by equipping language models with a concise selection of potential functions. Through the integration of Meta's \texttt{FAISS} for semantic search, we enhance the function description retrieval process, selecting the top 5 descriptions to accommodate the context length constraints of models such as Meta's Llama-7B and OpenAI's GPT-3.5.

Subsequently, we analyze the impact of training dataset size and model training methods on performance metrics.

\subsection{Android function calls}
To illustrate our model's application, we select Android system function calls as a case study, focusing on accuracy and latency in function call generation. Initially, we chose 20 Android APIs as our dataset foundation. We adopted two distinct methods for generating function call commands; for details on the API design, see the Appendix. The first method involves a RAG approach that identifies the most similar function descriptions based on the user query, which the language model then uses, along with the query, to generate the expected function call commands. We detail the various models employed in this evaluation below.

Utilizing Google Gemini, we sample relevant queries for the selected Android function calls and manually label the ground truth to form the evaluation dataset. We document our benchmark results, focusing on two critical metrics, accuracy and latency, as summarized in Table~\ref{tab:benchmark}.

\textbf{Llama-7B RAG Evaluation} \quad Initially, the pretrained Llama-7B model showed limited ability to generate the expected outcomes, leading us to employ a Llama-7B variant fine-tuned for function call generation \cite{llama-2-7b-chat-hf-function-calling-v3}. For the Llama-7B assessment, we applied the RAG method without strict output format requirements, considering responses with missing parentheses to be correct. This evaluation was conducted on a single NVIDIA A100 machine, with all Llama-7B results compared against the ground truth. The primary errors were incorrect function name selection and erroneous parameter generation. Despite employing few-shot learning to guide the model towards accurate function generation, the performance was modest, with an accuracy of 68.095\% when overlooking format requirements and a latency of 13.46 seconds, excluding model loading time. To improve latency, we implemented optimizations such as flash attention and a fast tokenizer.

\textbf{GPT-3.5 RAG Evaluation} \quad Similar to the approach with Llama-7B, we utilized GPT-3.5 for response generation, employing the same semantic search strategy for context acquisition. To enhance GPT-3.5's performance, we designed a specific prompt style, incorporating one-shot learning to improve accuracy further.

In this benchmark test, an impressive accuracy of 98.095\% was achieved, leveraging the \textit{gpt-3.5-turbo-0125} checkpoint, which is known for its optimization for function calling tasks. The latency was significantly improved to 1.97 seconds for generating a single function call, a notable enhancement over the Llama-7B model's performance. This improvement in speed is primarily attributed to the efficiency of the language model inference, as the RAG component remained consistent. GPT-3.5's quicker response may be due to OpenAI's use of multiple GPUs or a more advanced inference infrastructure. Further analysis revealed that a significant portion of the time was spent on content retrieval, despite only needing to fetch 5 function descriptions from a pool of 20. To optimize latency, the embeddings of all function descriptions were precomputed using OpenAI's \textit{text-embedding-3-large} endpoint, with \texttt{IndexFlatL2} employed for search indexing and parallel computation on multicore CPUs to enhance speed.

\begin{tcolorbox}
Below, you are presented with \{n\_candidates\} candidate functions. Your task is to analyze a specific query to determine which of these functions most appropriately addresses the query. Then, construct the correct function call with all necessary parameters, adhering to proper syntax.

Format for function call:

function\_name(param1, param2, ...)

Candidate Functions: \{candidates\}\\

def irrelevant\_function():

\quad'''If user query is not related to any of the predefined functions, this function will be called. Args: Returns:'''\\

Query:\{query\}

Example Scenario:

Query: "Change user's display mode to dark theme."

Given Functions: switch\_theme(theme), set\_brightness(level), irrelevant\_function()

Output: switch\_theme("dark")\\

Your goal is to select the most suitable function out of the \{n\_candidates\} candidates and generate an accurate function call that directly addresses the query. Ensure the output is a syntactically valid function call. Only return the function call.
\end{tcolorbox}

\textbf{GPT-3.5 and GPT-4 Evaluations} \quad To further reduce latency for GPT-3.5 and GPT-4, we included all 20 function descriptions directly in the context, bypassing the RAG method to avoid microservice interactions and their associated IO-bound overheads. This adjustment reduced latency to 1.18 seconds for GPT-3.5. The prompt template mirrored that of the GPT-3.5 RAG evaluation, with the addition of more candidate functions. However, accuracy slightly declined to 97.143\%, possibly due to diminished language model effectiveness with longer text inputs. Conversely, GPT-4 exhibited superior accuracy at 98.571\% and even lower latency than GPT-3.5, despite GPT-4 presumably being a larger model. This performance, evaluated on March 18 at 2 PM PDT, might reflect variances in API traffic or hardware configurations between the two models. GPT-4's enhanced performance suggests that OpenAI may allocate more GPU resources to it, or that it experiences less demand than GPT-3.5.

\textbf{Octopus model}\quad We now present the Octopus model, trained with 1000 data points sampled for each API, which achieves 99.524\% accuracy on our evaluation dataset. Moreover, the prompt used for this method is as simple as:

\begin{tcolorbox}
"Below is the query from the users, please call the correct function and generate the parameters to call the function. Query: \{user\_query\} Response:"
\end{tcolorbox}

In our approach, incorporating function information directly into the context is unnecessary, as the Octopus model has already learned to map functional tokens to the corresponding function descriptions, thereby conserving a significant number of tokens during processing. Given its compact size and the brevity of the required context, the Octopus model demonstrates a reduced latency of 0.38 seconds. To maintain an equitable comparison, we adhered to the same benchmark settings used for the Llama-7B evaluation, such as incorporating flash attention and not using quantization. Furthermore, we explored the deployment of our Octopus 2B model on mobile devices through quantization. By precomputing the state for the fixed prefix—\texttt{"Below is the query from the users, please call the correct function and generate the parameters to call the function. Query:"}—our on-device model achieves remarkable performance, completing a function call within 1.1 to 1.7 seconds for typical queries of 20 to 30 tokens using a standard Android phone.

\subsection{Extension to Vehicle, Yelp, and DoorDash function sets}
In addition to Android function calls, we expanded our evaluation to include 20 vehicle function calls, showcasing the algorithm's adaptability to diverse use cases. For vehicle functions, we focused on essential control methods such as volume adjustment, air conditioning, and seat positioning. We conducted benchmarks for vehicle functions paralleling the Android function evaluation and observed consistent performance patterns. Details on the vehicle functions are provided in the Appendix, enabling users to customize a new set of functional APIs for their specific needs. Furthermore, tests conducted with the Yelp and DoorDash APIs confirmed similar performance, underscoring our method's versatility across various function sets.

\subsection{Full and partial training datasets}
The Octopus model demonstrates exceptional performance when 1,000 data points are sampled for each API during its training phase. However, for training a new set of functions, cost efficiency becomes a consideration, given the need to generate a training dataset. In our analysis, generating 1,000 data points for a single API incurs a cost of 0.0224 USD, representing the investment required to train an Octopus-0 model for one specific function. By evaluating the Octopus-0, Octopus-2, and Octopus-3 models, we find that sampling only 100 data points per API can still achieve an accuracy of 98.095\%, as shown in Table~\ref{tab:benchmark}. Therefore, for individuals seeking to train their own Octopus model using our framework, we recommend a dataset size ranging from 100 to 1,000 data points per API.

\begin{table}[ht]
\centering
\setlength{\tabcolsep}{12pt} 
\renewcommand{\arraystretch}{1.} 
\begin{tabular}{@{}llp{7cm}@{}}
\toprule
\thead{Model name} & \thead{Training  Dataset Size} & \thead{Training configuration} \\
\midrule
Octopus-0 & 1K per API & Full model training \\
Octopus-1 & 1K per API & LoRA, rank = 16, lora\_alpha = 32, applied on \texttt{"q\_proj"}, \texttt{"k\_proj"}, \texttt{"v\_proj"}, \texttt{"o\_proj"}, \texttt{"up\_proj"}, \texttt{"down\_proj"} \\
Octopus-2 & 500 per API & Full model training \\
Octopus-3 & 100 per API & Full model training \\
\bottomrule
\vspace{1pt}
\end{tabular}
\caption{Configurations of the four different Octopus models.}
\label{tab:octopus_models}
\end{table}

\subsection{Full training and LoRA training}
LoRA plays a crucial role in our framework, particularly when integrating the Octopus model across multiple applications to ensure smooth computation. Instead of employing full models for each API set, we opt for diverse LoRA trainings tailored to the specific function setups of different apps. As Table~\ref{tab:benchmark} shows, switching to LoRA training results in a minor accuracy decrease. Nonetheless, the maintained accuracy remains sufficiently high for production deployment.

\subsection{Parallel and nested function calls}
The benchmark tests above are intended for single function calls. To enable parallel and nested function calls, we need to prepare 4K data points for each API so that the accuracy can reach the same level as for single function calls.

\subsection{Weighted loss function for special tokens}
A distinctive aspect of our approach involves incorporating numerous special tokens into the tokenizer and expanding the language model's head. The loss function is defined as follows:

\begin{equation}\label{equ:loss}
\mathcal{L} = -\sum_{t=1}^{T} \sum_{i}^V y_{t,i} \log(\hat{y}_{t,i}),
\end{equation}
where $T$ represents the sequence length, and $V$ denotes the vocabulary size.

Given the introduction of special tokens ranging from \texttt{<nexa\_0>} to \texttt{<nexa\_N-1>}, along with the distinct token \texttt{<nexa\_end>}, which are absent from the Gemma-2B pretraining dataset, we confront an imbalanced dataset challenge during model training. To address this, we adopt a weighted cross-entropy loss as a surrogate loss to improve convergence:
\begin{equation}
\mathcal{L} = -\sum_{t=1}^{T} \sum_{i}^V\omega_{i} y_{t,i} \log(\hat{y}_{t,i}).
\end{equation}

\begin{figure}[h]
    \centering
    \includegraphics[width=0.78\textwidth]{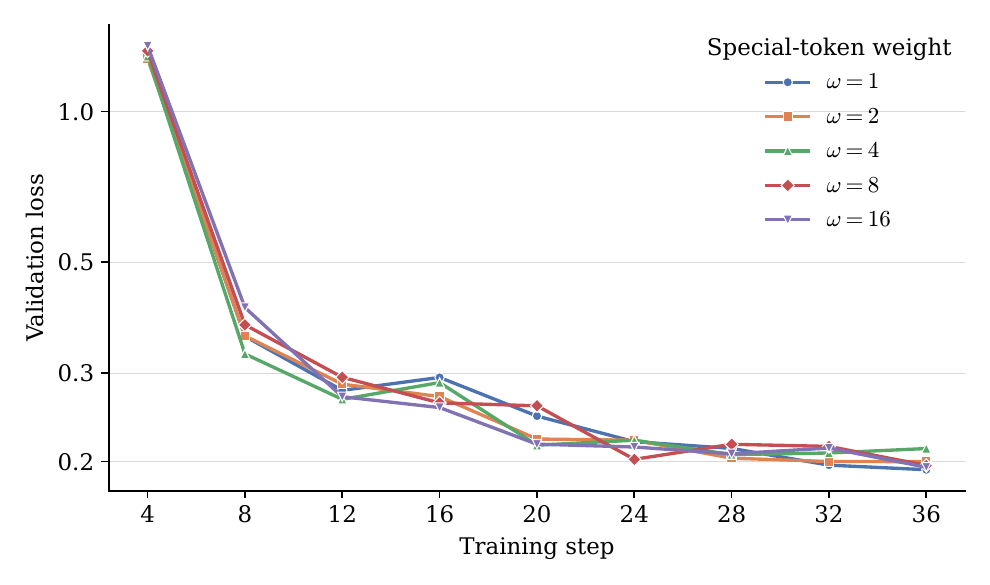}
    \caption{Validation loss, evaluated with the unweighted objective of Equation~(\ref{equ:loss}), for models trained with weighted surrogate losses using special-token weights $\omega \in \{1, 2, 4, 8, 16\}$. All weight settings converge to comparable validation loss.}
    \label{fig:token_weight}
\end{figure}

In our configuration, non-special tokens are assigned a weight of 1, while special tokens receive elevated weights. Early-stage training experiments indicate that increasing the token weight can expedite convergence. The validation loss, based on Equation (\ref{equ:loss}) with varying surrogate losses used for training, is illustrated in Figure (\ref{fig:token_weight}). Our findings suggest that employing a surrogate training loss early in the training process aids convergence. Nonetheless, experiments reveal no performance disparity in the fine-tuned model, nor significant differences in wall-clock time. Therefore, utilizing an equal-weighted token loss is recommended when the number of functional tokens is small. In our benchmark tests, the evaluated model is trained with equal token weights.

\section{Discussion and future work}
Our current training initiative demonstrates that any specific function can be encapsulated in a newly coined term, the \textit{functional token}, a novel token type seamlessly integrated into both the tokenizer and the model. Through a cost-effective training process amounting to merely two cents, this model facilitates the deployment of AI agents characterized by remarkably low latency and high accuracy.

The potential impacts of our research are extensive. For application developers, including those at DoorDash and Yelp, our model paves the way for training on application-specific scenarios. Developers can pinpoint the APIs most used by their audience, transform them into functional tokens for the Octopus model, and proceed with deployment. This strategy has the capacity to fully automate app workflows, emulating functionality akin to Apple's Siri, albeit with significantly faster response speeds and higher accuracy.

Furthermore, the model's application within the operating systems of PCs, smartphones, and wearable technology presents another exciting avenue. Software developers could train small LoRAs specific to the operating system. By accumulating multiple LoRAs, the model facilitates efficient function calling across diverse system components. For instance, incorporating this model into the Android ecosystem would enable Yelp and DoorDash developers to train distinct LoRAs, thus rendering the model operational on mobile platforms as well.

Looking ahead, we aim to develop a model dedicated to on-device reasoning. Our ambitions are twofold: first, to achieve notable speed enhancements for cloud deployments, vastly outpacing GPT-4 in speed metrics; and second, to support local deployment, offering a valuable solution for users mindful of privacy or operational costs. This dual deployment strategy not only extends the model's utility across cloud and local environments but also caters to user preferences for either speed and efficiency or privacy and cost savings.

\newpage
\medskip
{\small
\bibliographystyle{plainnat}
\bibliography{citation}
}


\section*{Appendix}
\subsection*{A.1 Android function examples}
\begin{lstlisting}[style=pythonstyle]
def take_a_photo(camera="back", resolution="1080p"):
    """
    Captures a photo using the specified camera and resolution settings.

    Parameters:
    - camera (str, optional): Specifies the camera to use. Can be 'front' or 'back'. The default is 'back'. Optional to provide.
    - resolution (str, optional): Sets the photo resolution. Options include '720p', '1080p', and '4K'. The default is '1080p'. Optional to provide.

    Returns:
    - str: The string contains the file path of the captured photo if successful, or an error message if not. Example: '/storage/emulated/0/Pictures/MyApp/IMG_20240310_123456.jpg'
    """


def get_trending_news(category=None, region='US', language='en', max_results=5):
    """
    Fetches trending news articles based on category, region, and language.

    Parameters:
    - category (str, optional): News category to filter by, by default use None for all categories. Optional to provide.
    - region (str, optional): ISO 3166-1 alpha-2 country code for region-specific news, by default, uses 'US'. Optional to provide.
    - language (str, optional): ISO 639-1 language code for article language, by default uses 'en'. Optional to provide.
    - max_results (int, optional): Maximum number of articles to return, by default, uses 5. Optional to provide.

    Returns:
    - list[str]: A list of strings, each representing an article. Each string contains the article's heading and URL.
    """


def get_weather_forecast(location, days=1):
    """
    Provides a weather forecast for a specified location over a given number of days. Each day's forecast includes a brief description of the expected weather conditions.

    Parameters:
    - location (str): The location for which the weather forecast is desired. Can be a city name, ZIP code, or other location identifiers.
    - days (int, optional): The number of days to include in the forecast, starting from today. The default is 1 day. Optional to provide.

    Returns:
    - list[str]: A list of strings, each representing the weather forecast for one day. Each string includes the date and a brief description of the weather conditions. Formatted in 'YYYY-MM-DD: Description' format.
    """


def send_email(recipient, subject, body, attachments=None, cc=None, bcc=None):
    """
    Sends an email with optional attachments, CC, and BCC.

    Parameters:
    - recipient (str): Primary recipient's email address.
    - subject (str): Email subject line.
    - body (str): Main email body content.
    - attachments (list of str, optional): A list of file paths representing files to attach to the email. Defaults to None, indicating no attachments. Optional to provide.
    - cc (list of str, optional): A list of email addresses to include in the Carbon Copy (CC) field. Defaults to None. Optional to provide.
    - bcc (list of str, optional): A list of email addresses to include in the Blind Carbon Copy (BCC) field. Defaults to None. Optional to provide.

    Returns:
    """


def search_youtube_videos(query, max_results=10, search_filter="Relevance"):
    """
    Searches YouTube for videos matching a query.

    Parameters:
    - query (str): Search query.
    - max_results (int, optional): Maximum number of search results, by default, use 10. Optional to provide.
    - search_filter (enum, optional): Filter for search results, chosen from 'Relevance', 'Upload date', 'View Count', 'Rating'. By default, use 'Relevance'. Optional to provide.

    Returns:
    - list[str]: A list of strings, each string includes video names and URLs.
    """
\end{lstlisting}

\subsection*{A.2 Vehicle function examples}
\begin{lstlisting}[style=pythonstyle]
def adjust_volume(volume_diff=None, set_value=None):
    """
    Adjusts the device's volume by a specified difference or sets it to a specified value. Only one operation can be performed at a time.

    Parameters:
    - volume_diff (int, optional): The amount to adjust the current volume by. Positive to increase, negative to decrease, optional to provide.
    - set_value (int, optional): The target volume level to set, in the range of 0 to 50, optional to provide.

    Note:
    - If both `volume_diff` and `set_value` are provided, only one will be considered based on the implementation's logic.
    
    Returns:
    - bool: True if the volume was adjusted successfully, False otherwise.
    """

def set_climate_temperature(zone, temperature):
    """
    Configures the temperature for a specific zone within the vehicle's climate control system.

    Parameters:
    - zone (str): The zone to set the temperature for ('driver', 'passenger', 'rear').
    - temperature (int): The target temperature in Fahrenheit, within the range of 60 to 80 degrees.
    
    Returns:
    - bool: True if the temperature was set successfully, False otherwise.
    """

def adjust_seat_position(seat, position, distance):
    """
    Modifies the position of a specified seat by a certain distance.

    Parameters:
    - seat (str): The seat identifier ('driver', 'passenger').
    - position (str): The direction to adjust the seat in ('forward', 'backward', 'up', 'down').
    - distance (int): The amount of adjustment in millimeters.

    Returns:
    - bool: True if the seat was adjusted successfully, False otherwise.
    """

def control_window(window, position, distance):
    """
    Adjusts a vehicle window's position by a specific distance.

    Parameters:
    - window (str): The window to control ('front left', 'front right', 'rear left', 'rear right').
    - position (str): The direction to move the window ('up' or 'down').
    - distance (int): The distance to move the window, in millimeters.

    Returns:
    - bool: True if the window was adjusted successfully, False otherwise.
    """

def operate_sunroof(action, intensity=None):
    """
    Operates the sunroof with a specified action and optional intensity.

    Parameters:
    - action (str): The sunroof operation to perform ('open', 'close', 'tilt').
    - intensity (int, optional): The degree to which the sunroof should be opened or tilted, as a percentage, optional to provide.

    Returns:
    - bool: True if the sunroof was operated successfully, False otherwise.
    """
\end{lstlisting}
\end{document}